\pdfoutput=1

\documentclass[11pt]{article}

\usepackage[final]{acl}

\usepackage{times}
\usepackage{latexsym}

\usepackage[T1]{fontenc}

\usepackage[utf8]{inputenc}
\usepackage{amsmath}
\usepackage{microtype}

\usepackage{inconsolata}

\usepackage{makecell}
\usepackage{cellspace}
\usepackage{stfloats}
\usepackage{longtable}
\usepackage{booktabs}
\usepackage{tikz, xcolor}
\usepackage[most]{tcolorbox}
\usepackage{array,multirow,multicol,graphicx,subcaption}
\usepackage[most]{tcolorbox} 

\definecolor{customblue}{HTML}{2E86C1}
\definecolor{customgreen}{HTML}{28B463}
\definecolor{custombrown}{HTML}{CA6F1E}

\title{OCR or Not? Rethinking Document Information Extraction in the MLLMs Era with Real-World Large-Scale Datasets}

\author{Jiyuan Shen$^{\textbf{1}}$, Peiyue Yuan$^{\textbf{1}}$, Atin Ghosh$^{\textbf{1}}$, Yifan Mai$^{\textbf{2}}$, Daniel Dahlmeier$^{\textbf{1}}$ \\
  $^{1}$SAP \quad $^{2}$Stanford University \\
  \small{\texttt{\{jiyuan.shen,peiyue.yuan,atin.ghosh,d.dahlmeier\}@sap.com} \quad \texttt{yifan@cs.stanford.edu}}
}

\begin{document}
\maketitle
\begin{abstract}
Multimodal Large Language Models (MLLMs) enhance the potential of natural language processing. However, their actual impact on document information extraction remains unclear. In particular, it is unclear whether an MLLM-only pipeline—while simpler—can truly match the performance of traditional OCR+MLLM setups. In this paper, we conduct a large-scale benchmarking study that evaluates various out-of-the-box MLLMs on business-document information extraction. To examine and explore failure modes, we propose an automated hierarchical error analysis framework that leverages large language models (LLMs) to diagnose error patterns systematically. Our findings suggest that OCR may not be necessary for powerful MLLMs, as image-only input can achieve comparable performance to OCR-enhanced approaches. Moreover, we demonstrate that carefully designed schema, exemplars, and instructions can further enhance MLLMs performance. We hope this work can offer practical guidance and valuable insight for advancing document information extraction.  

\end{abstract}

\section{Introduction}

Within the field of natural language processing (NLP), a key application involves automatically extracting key information from various sources, such as invoices, insurance quotes, and financial statements, and turning it into structured information. This capability is used in various industries, which help businesses automate and streamline document-based and scene-text workflows, improving operational efficiency \cite{gartner-reviews}.

However, the vast majority of mature document information extraction systems in the industry still rely on a two-stage framework, where optical character recognition (OCR) first extracts textual content before a secondary specialized model converts the text into structured information following a schema \cite{wang2023vrdu}. This approach, while effective, is inherently complex, difficult to generalize to new domains and susceptible to error propagation from OCR to downstream extraction. These limitations have motivated growing interest in OCR-free and few-shot learning approaches \cite{kim2022ocr, ye2023mplug, liu2024textmonkey, mistralocr}. The rapid advancement of general-purpose MLLMs further strengthens this trend, as many are pretrained on large-scale structured document and should, in principle, possess strong information extraction capabilities \cite{team2024gemini, intelligence2024amazon}. Yet their true effectiveness in this area remains highly unclear.

Therefore, we evaluate a range of state-of-the-art MLLMs on a large-scale, high-quality benchmark dataset, which reflects our experience in developing enterprise document AI services. Specifically, we experiment with three different input modalities: OCR-extracted text only, raw document images only, and a combination of both. 

Furthermore, we leverage large language models (LLMs) capabilities to develop an automated error analysis framework that systematically categorizes prediction errors through a hierarchical reasoning approach. By analyzing failure cases and benchmarking results, we provide deeper insights into critical questions, such as \textit{Is OCR necessary for MLLM-based document information extraction? Can MLLMs serve as a promising path for streamlining the pipeline?} Through this study, our objective is to bridge the gap between academic research and real-world applications, shedding light on the strengths and limitations of advanced approaches in document information extraction.

The main contributions of this work are summarized as follows:  
\begin{enumerate}  
    \item We investigate the role of OCR in document information extraction with MLLMs and find that for specific powerful models, OCR may not be necessary and can even have a slightly negative impact. Our findings suggest that MLLM-only pipeline is a promising direction for document information extraction.  
    \item We demonstrate that as MLLMs increase in size, their information extraction performance can still improve accordingly.
    \item We propose a hierarchical error analysis framework that can automatically discover the error patterns.
    \item We find that general-purpose MLLMs lack task-specific knowledge, highlighting the need for more carefully designed schemas, exemplars, and instructions. We refine our approach and achieve measurable performance improvement by leveraging insights from our error analysis framework.
\end{enumerate}

\section{Related Work}

Although using domain-specific OCR models together with task-tuned extraction models is widely regarded as good practice in industry \cite{katti2018chargrid, huang2022layoutlmv3}, the drawbacks are easy to recognize: system complexity, limited generalization, and substantial labor required to adapt pipelines to new domains. These limitations have motivated the research community to explore more streamlined end-to-end approaches, even at the cost of a slight performance trade-off~\cite{ouyang2025omnidocbench}. The rapid advancement of MLLMs has further accelerated this shift \cite{mistralocr, bai2023qwen}. These powerful models are pretrained on large-scale, diverse image datasets and subsequently refined through instruction tuning, enabling strong visual understanding, layout awareness, and zero-shot reasoning. For example, GPT-4o \cite{hurst2024gpt} and Gemini \cite{team2023gemini} exhibit impressive capabilities in jointly interpreting visual layouts and textual content, offering a promising balance between accuracy and efficiency. However, a comprehensive benchmark of MLLMs for business-document information extraction is still lacking. To address this gap, we aim to provide a rigorous evaluation and a fair comparison of their effectiveness in real-world scenarios.

\begin{figure}[]
    \centering 
    \includegraphics[width=0.95\linewidth]{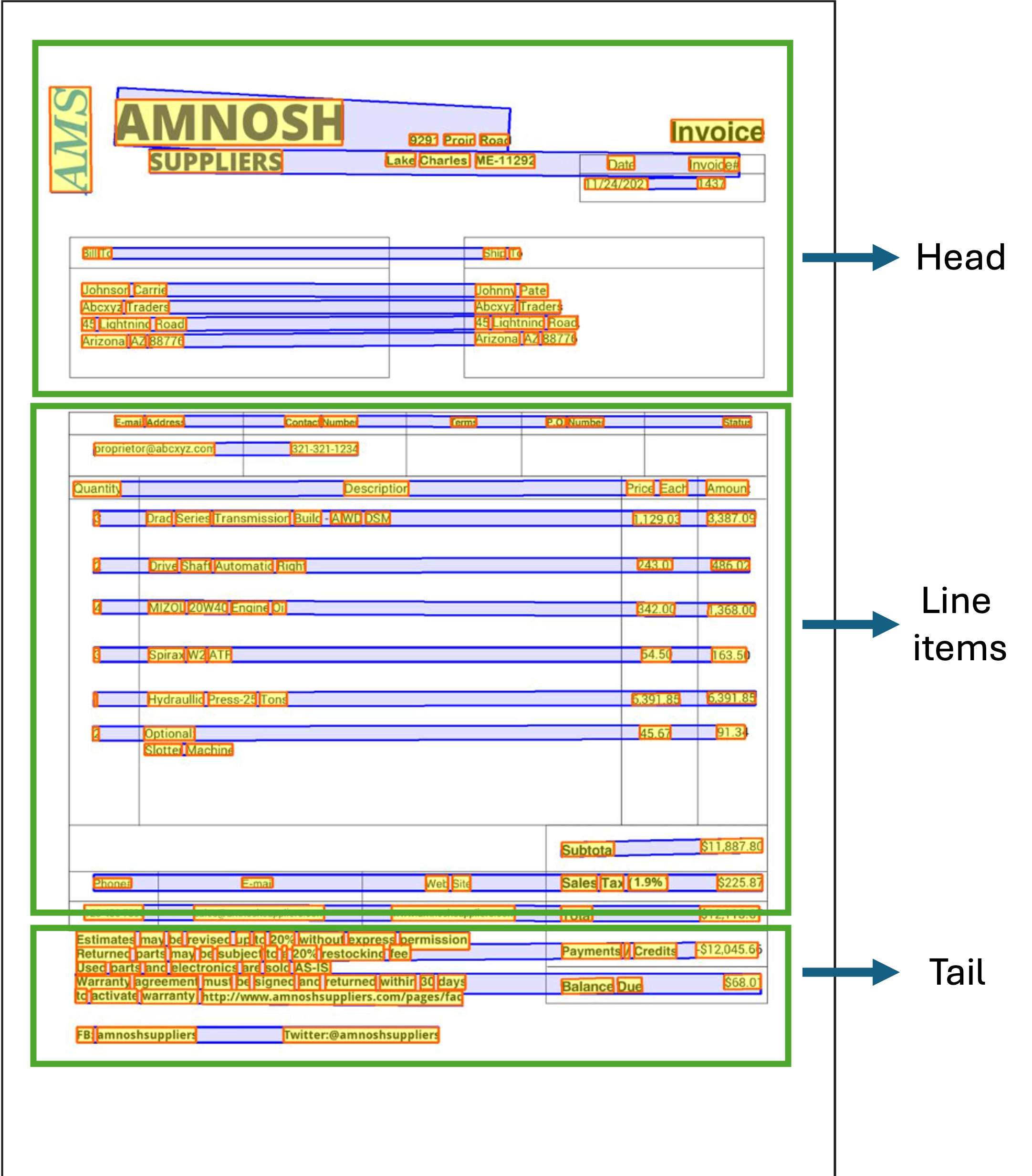}
    \caption{Example of a document page extracted using our OCR engine.}
    \label{fig:ocr_example}
\end{figure}

\section{Methodology}
\subsection{Internal Industrial Document Dataset}
\begin{figure}[]
    \centering 
    \includegraphics[width=0.95\linewidth]{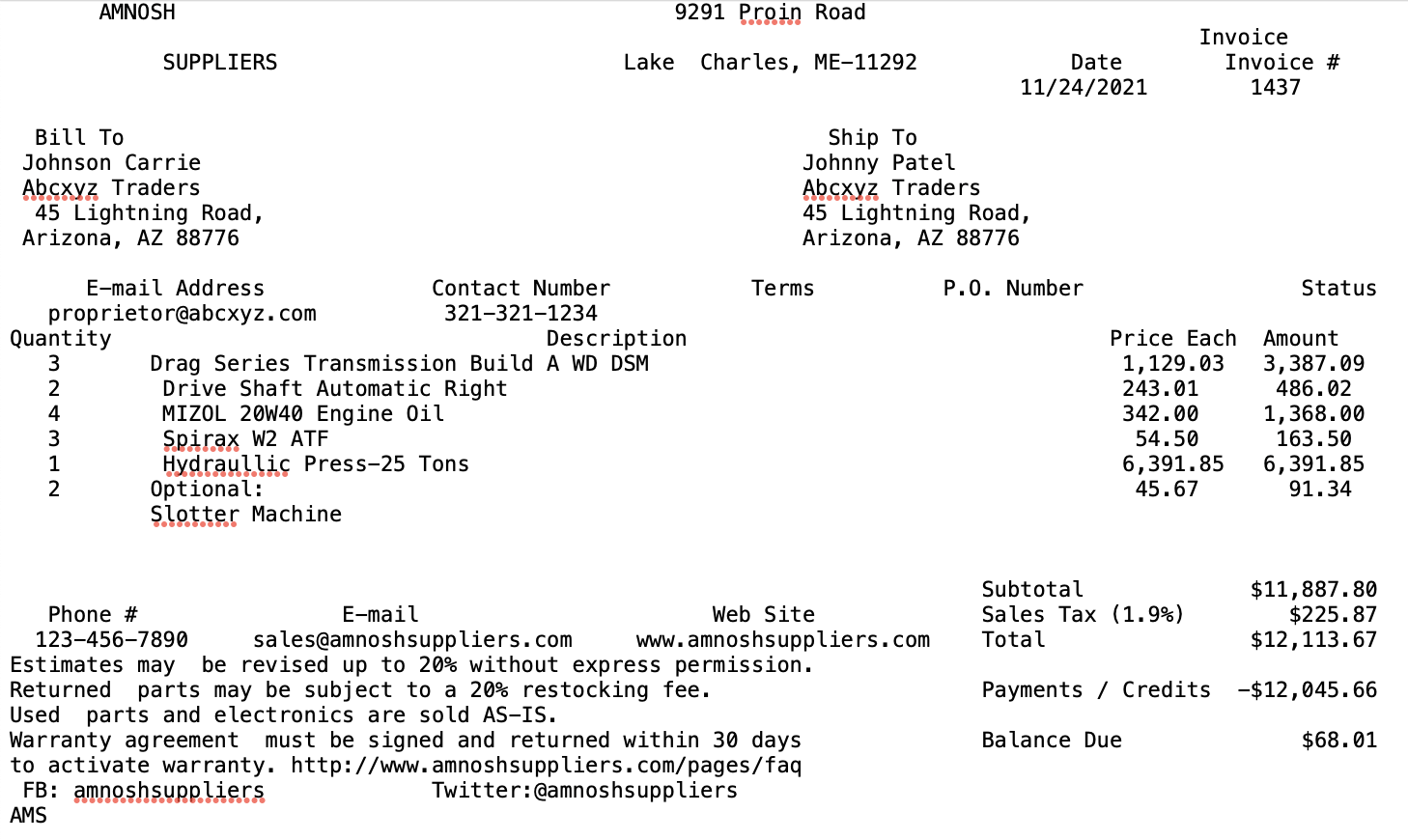}
    \caption{An example of textual content extracted by our in-house OCR engine.}
    \label{fig:ocr_result}
\end{figure}

Our internal datasets encompass a diverse range of documents, with dataset C1 sourced from the supply chain domain and C2 from finance. 
For all of these documents, we collected manual annotations with carefully curated structured ground-truth labels, along with OCR-extracted text results. 
We use our in-house OCR engine that has been developed for business documents and achieves high performance with an average accuracy of more than 90\% in multiple languages. In our internal evaluations, it outperforms state-of-the-art OCR methods and OCR services provided by major machine learning platforms. Figure~\ref{fig:ocr_example} shows a sample document page, and Figure~\ref{fig:ocr_result} illustrates the textual content extracted by our OCR engine. As demonstrated, we preserve layout information by retaining whitespace as a structural delimiter in the extracted text.

Compared with existing open-source datasets, ours is substantially more challenging. The difficulties stem primarily from two sources: (i) multilingual content and (ii) structural complexity. Regarding multilinguality, we provide comprehensive statistics analysis in Appendix~\ref{appendix:data distribution} that reflect the wide language distribution across multiple countries and multi-page documents. Regarding structural complexity, our dataset contains nested information, stacked cells within line items, and heterogeneous header structures—factors that significantly increase the difficulty of document parsing. Refer to Figure~\ref{fig:ocr_example} for an example.

\subsection{Evaluation Pipeline and Metrics}

We have incorporated some of the principles of VHELM's design and utilize wrapped clients \cite{lee2024vhelm}. Our evaluation pipeline consists of three main stages. The first stage involves using an OCR engine to extract textual content from document images, preserving the positional information. For image-only experiments, the OCR step is skipped. 

The second stage focuses on structured information extraction. For MLLM-based approaches, we construct a prompt template (see Appendix \ref{details in eval} for details) that includes format instructions and the document schema, enabling zero-shot information extraction. The target extraction schema consists of \emph{header fields} and a list of \emph{line items}, which capture structured tabular information. The MLLM output is a JSON object, where keys represent entity types, and values correspond to extracted content from the document. An example of response is shown in Appendix \ref{details in eval}.

In the final stage, we report the overall performance using the standard F1 score. Specifically, since our outputs are structured as key–value pairs, we compute precision and recall over all key–value predictions, and then derive the F1 score from these metrics.

\subsection{Hierarchical Error Analysis Framework}

\begin{figure}[t]
    \centering 
    \includegraphics[width=\linewidth]{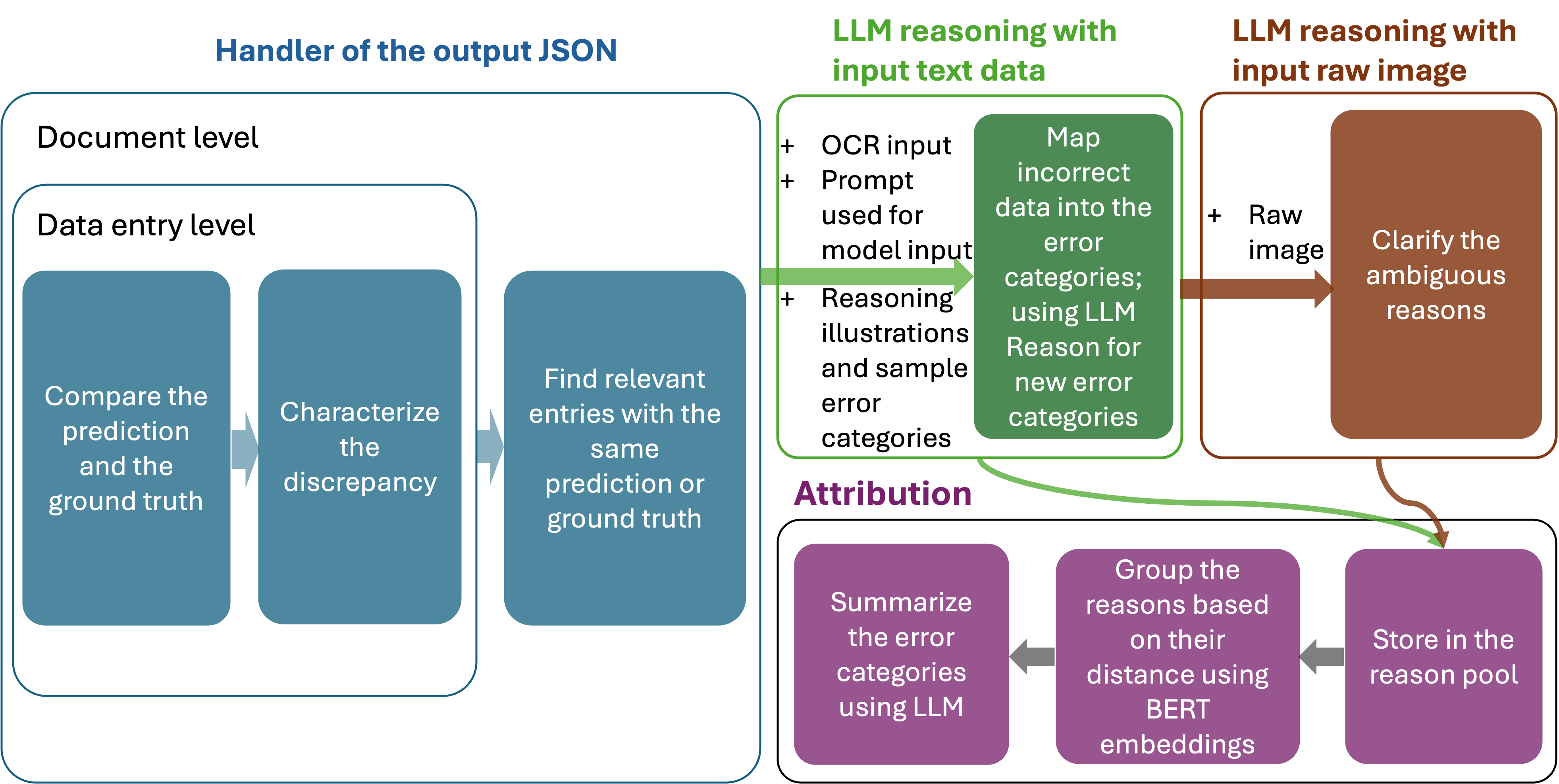}
    \caption{Hierarchical Error Analysis Framework}
    \label{fig:error_analysis_framework}
\end{figure}

To systematically diagnose errors in document information extraction, we adopt a hierarchical error analysis framework inspired by \citet{chen2024automatic}. Our framework categorizes errors from the middle to the highest level, following a logical progression from direct observations to deeper root causes. This structured approach ensures that errors are first identified based on surface-level discrepancies and then further analyzed to uncover underlying reasons. We show our framework in Figure \ref{fig:error_analysis_framework}.

\subsubsection{Handler}
The error analysis process begins with an automated error handler that systematically logs and classifies prediction mismatches. Given a set of extracted predictions and ground-truth values, we compare them at both character and semantic levels, ensuring robust error identification. The analysis is performed at both the field level and document level. The process consists of three main steps: (1) comparing the predicted values with the ground truth, (2) characterizing the discrepancies between them, and (3) identifying relevant entries with similar predictions or ground-truth values for further analysis.

\subsubsection{LLM Reasoning}
To refine the classification of errors and the root cause analysis, we use LLM-based reasoning. Instead of manually analyzing failure cases, we employ LLMs and MLLMs to help generate structured diagnostic reports. 

The hierarchical reasoning process consists of two steps: (1) mapping incorrect predictions into predefined error categories using LLMs, which also allows for identifying new error categories when necessary, and (2) clarifying ambiguous errors by incorporating raw document images as additional input for reasoning. The first step utilizes textual input from OCR results, predicted values, and ground-truth labels, along with predefined reasoning templates and few-shot cause-of-failure examples to categorize errors and generate potential causes. In cases where textual reasoning alone is insufficient, such as errors arising from layout complexities or visual ambiguities, we introduce raw document images to refine error attribution. This approach ensures a more comprehensive understanding of extraction failures. By the end of this stage, all errors are categorized into mid-level error reasons, which form a structured foundation for deeper analysis in subsequent attribution steps.

\subsubsection{Attribution}
The final stage of our framework involves attributing errors to specific highest-level failure sources. Post-processing is performed on the LLM-generated explanations to summarize the error categories. First, the categorized reasons are stored in a structured reason pool. Next, we apply BERT-based embedding clustering to group similar reasons based on cosine similarity, ensuring a coherent categorization of error types. Finally, we extract representative keywords for each error type within the same cluster.
We analyze the behavior of the model across multiple documents to determine whether errors originate from OCR misrecognition, layout misinterpretation, prompt misalignment, model capability issues, or schema inconsistencies.

\section{Experiments}

\subsection{Baselines}

We evaluate each MLLM using three input formats: document image-only, OCR-extracted text, and a combination of both. Our experiments focus on flagship models from major providers, limited to those released after 2024 to reflect state-of-the-art capabilities. Gemini 2.5 Flash is used in place of the Gemini 2.5 Pro due to Gemini 2.5 Pro's high inference latency.

Although current open-source models are generally still underperform in comparison to proprietary models, we add Llama 4 for a comprehensive benchmarking.

\begin{table*}[]
    \centering
    \renewcommand{\arraystretch}{1.04}
    \setlength{\tabcolsep}{5pt}
    \caption{Performance comparison of different MLLMs across evaluation settings: Image, OCR, and Image + OCR as input formats. C1 and C2 refer to two different datasets, while Mean denotes the arithmetic mean of the F1-scores on C1 and C2.}
    \label{tab:exp_results}
    \resizebox{\textwidth}{!}{
    \begin{tabular}{ccccc|ccc|ccc}
        \toprule
        \multirow{2}{*}{\textbf{Company}} & \multirow{2}{*}{\textbf{Model}} & \multicolumn{3}{c|}{\textbf{Image-only}} & \multicolumn{3}{c|}{\textbf{OCR-only}} & \multicolumn{3}{c}{\textbf{Image + OCR}} \\
        \cmidrule(lr){3-5} \cmidrule(lr){6-8} \cmidrule(lr){9-11}
        & & Dataset C1 & Dataset C2 & Mean & Dataset C1 & Dataset C2 & Mean & Dataset C1 & Dataset C2 & Mean \\
        \midrule
        \multirow{2}{*}{Meta} & Llama 4 Scout  & 67.4                        & 69.3       & 68.4      & 68.1    & 69.7       & 68.9    & 67.3   & 69.8  & 68.6  \\
        & Llama 4 Maverick  & 62.8 & 68.2 & 65.5  & 63.9 & 68.1         & 66.0 & 62.9    & 68.2   & 65.5  \\
        \midrule
        MistralAI & Pixtral Large (2411) & 68.7 & 57.4 & 63.1 & 75.3 & 71.2 & 73.3 & 72.7 & 68.0 & 70.4 \\
        \midrule
        Amazon & Nova Pro & 77.9 & 65.1 & 71.5 & 68.7 & 65.1 & 66.9 & 77.5 & 66.6 & 72.1 \\
        \midrule
        \multirow{2}{*}{OpenAI} & GPT-4o mini & 68.3 & 64.9 & 66.6  & 66.1 & 70.5 & 68.3 & 71.6 & 70.5 & 71.1 \\
                                & GPT-4o & 75.5 & 68.9 & 70.1  & 76.0 & 69.5 & 72.8 & 76.7 & 69.3 & 73.0 \\
                           
        \midrule
        \multirow{2}{*}{Anthropic} & Claude 3 Opus & 43.8 & 56.4 & 50.1 & 72.0 & 68.2 & 70.1 & 74.0 & 69.1 & 71.5 \\
                                   & Claude 3.5 Sonnet & 65.0 & 69.3 & 67.2 & 73.7 & \textbf{72.6} & 72.8 & 73.6 & 69.6 & 71.6 \\
        \midrule
        \multirow{3}{*}{Google} & Gemini 1.5 Pro & \textbf{87.3} & 66.4 & \textbf{76.8} & \textbf{78.4} & 69.8 & \textbf{74.1} & \textbf{86.2} & 65.0 & \textbf{75.6} \\
                                   & Gemini 2.0 Pro & 75.2 & \textbf{73.3} & 74.3 & 77.6 & 69.5 & 73.6 & 77.1 & \textbf{73.2} & 75.2 \\
                                   & Gemini 2.5 Flash & 73.9 & 71.2 & 72.6 & 74.6 & 69.6 & 72.1 & 73.0 & 71.4 & 72.2 \\
        
        \bottomrule
    \end{tabular}
    }
    \label{Large_llm_main_results}
\end{table*}

\subsection{Experiment Results}
\label{exp_results}

Table \ref{Large_llm_main_results} presents a comparative analysis of various MLLMs under three input settings: image-only, OCR-only, and image + OCR. Model performance is evaluated using the F1-score on two business document datasets—C1 (from the supply chain domain) and C2 (from the finance domain)—with the arithmetic mean used as the overall metric.

Models that accept OCR-only input consistently achieve mean F1-scores in the range of 66\% to 74\%, exhibiting relatively low variance across the board. In contrast, image-only inputs result in a wider performance spread, highlighting larger disparities among models from different providers. Notably, when OCR and image inputs are combined, the variance in mean performance decreases, with scores falling within a narrower range of 70\% to 75\%. This suggests that incorporating image input can help models produce more stable and robust predictions.

A row-level comparison with the OCR-only setting further reveals that models such as Nova Pro, GPT-4o, and the Gemini series benefit from multimodal input, which shows improvements of 1–3 percentage points in F1-score. However, exceptions do exist. For example, models like Pixtral and Claude 3.5 Sonnet exhibit decreased performance when image input is added. We hypothesize that these models may struggle to effectively integrate visual information with their text processing components, leading to suboptimal fusion of multimodal features.

\subsection{Analysis}

\subsubsection{Is OCR necessary for MLLM-based document information extraction?}

From Table~\ref{tab:exp_results}, we observe an interesting phenomenon when analyzing the flagship Gemini and Nova models. Unlike several other models, the performance of these two model series does not significantly degrade when using image-only input, without OCR-extracted text. In some cases, they even exhibit notable improvements. While this was initially considered a potential anomaly, the trend remained consistent across multiple re-evaluations with varied sampling strategies. This implies that certain advanced multimodal models are capable of directly extracting structured information from document images and comprehending textual content effectively, without the need for OCR as an intermediary. In particular, for the Gemini models, OCR-generated text appears to provide little to no additional benefit. We provide more explanation in Appendix~\ref{appendix:explanation}.

\subsubsection{Does MLLMs performance scale with model size across different input modalities?}

\begin{figure}[t]
    \centering 
    \includegraphics[width=\linewidth]{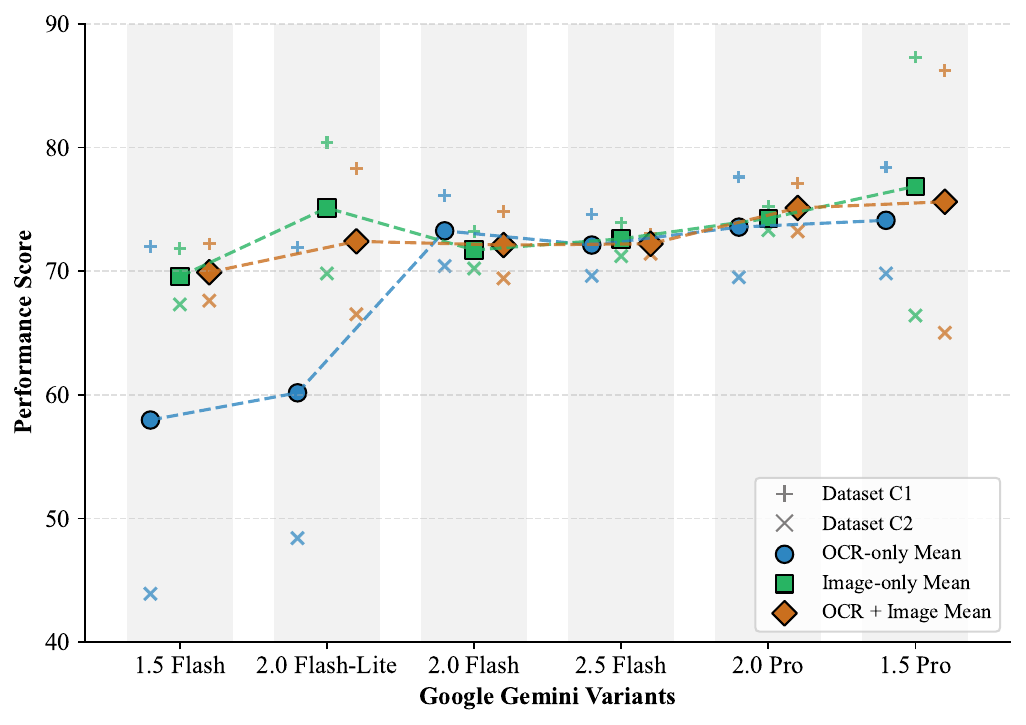}
    \caption{Performance comparison on various size models across different input types. The small shape (\protect\raisebox{-1pt}{\protect\tikz\protect\fill[customblue] (0,0) circle (3pt);}, 
    \protect\raisebox{-1pt}{\protect\tikz\protect\fill[customgreen] (0,0) rectangle (0.2,0.2);},  
    \protect\raisebox{-1pt}{\protect\tikz\protect\fill[custombrown] (0,0) -- (0.1,0.1) -- (0,0.2) -- (-0.1,0.1) -- cycle;}) 
    denotes the arithmetic mean across two different categories of dataset. $+$ is the F1-score in C1, while $\times$ is for C2.}
    \label{fig:scaling_perf}
\end{figure}

It is well established that larger models will perform better \cite{kaplan2020scaling}. However, does this trend persist within our internal dataset when using different input modalities for MLLMs? Specifically, as shown in Figure \ref{fig:scaling_perf}, the overall performance improves as the size of the model increases\footnote{Google does not disclose the exact parameter sizes for each variant, but the size relationship can be partially inferred from the model naming.}. Among the three input types, the most significant performance gain is observed with OCR-only input, where the score increases from 57\% to 74\%. In contrast, the performance of image-only and multimodal inputs remains relatively comparable. The potential reason is that even the Gemini 1.5 Flash is already capable of a rather high baseline performance score.

A particularly interesting observation is that for the Gemini 2.0 Flash-Lite model, the image-only input outperforms the multimodal input by nearly 3\%. This result suggests that OCR-extracted text does not necessarily provide a significant performance boost. Instead, even the powerful, yet small model can extract and understand textual information directly from images without relying on explicit OCR input. Furthermore, the variance in performance across modalities suggests that different model sizes exhibit varying levels of dependence on OCR-extracted text. Meanwhile, interestingly, for open-source MLLMs such as the Llama 4 series, we observe a negative correlation between model size and multimodal performance gains. This may stem from differences in training corpus scale—for instance, the smaller Scout model is trained on 40T tokens, whereas the larger Maverick model uses only 22T tokens—potentially limiting the larger model’s OCR robustness and cross-modal alignment.

Taken together, these findings offer new insights into MLLM scaling behavior and highlight the substantial potential of vision encoders to handle textual information effectively, especially when using genuinely high-capacity MLLMs.

\subsubsection{Computational cost and inference latency}

Since most of the models we benchmark are closed-source, we report the average cost and inference latency by directly consuming these endpoint. From Table \ref{cost_table}, we observe that both speed and cost continue to improve over time. Additionally, MLLMs offer a key advantage — their strong generalization capability. They can adapt more easily to new document types and languages without requiring extensive task-specific fine-tuning. This brings us back to our core motivation: MLLMs hold significant potential to streamline the entire document processing pipeline while maintaining strong performance in information retrieval tasks.

\begin{table}[h!]
\centering
\small
\setlength{\tabcolsep}{4pt} 
\caption{Estimated latency and cost per page for different closed-source models.}
\begin{tabular}{lcc}
\toprule
\textbf{Model} & \textbf{Latency/Page (Est.)} & \textbf{Cost/Page (Est.)} \\
\midrule
GPT-4o & $\sim$2.2s & $\sim$\$0.006 \\
Claude 3.5 Sonnet & $\sim$4.7s & $\sim$\$0.010 \\
Claude 3 Opus & $\sim$7.0s & $\sim$\$0.050 \\
Gemini 1.5 Pro & $\sim$3.9s & $\sim$\$0.001 \\
Gemini 2.0 Pro & $\sim$2.0s & $\sim$\$0.004 \\
Gemini 2.5 Flash & $\sim$1.4s & $\sim$\$0.0025 \\
Pixtral Large & $\sim$7.0s & $\sim$\$0.0035 \\
Amazon Nova Pro & $\sim$6.6s & $\sim$\$0.004 \\
\bottomrule
\end{tabular}
\label{cost_table}
\end{table}

\section{Discussion}
\label{discussion}

\begin{figure}[t]
    \centering 
    \includegraphics[width=\linewidth]{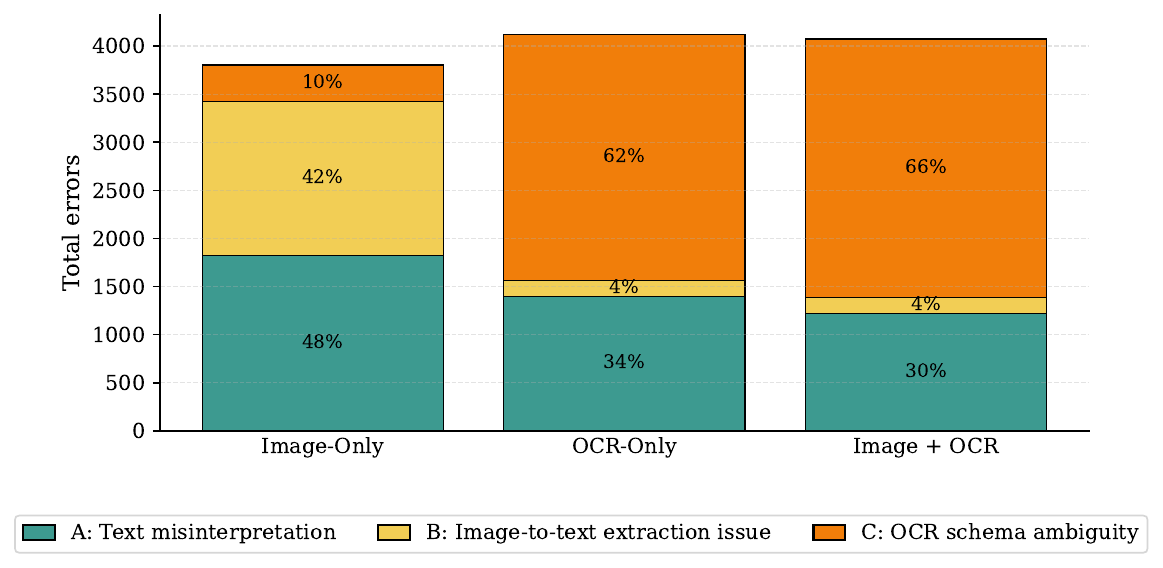}
    \caption{Error analysis results for three different input modalities.}
    \label{fig:error_pie_chart}
\end{figure}

We employ our hierarchical error analysis framework to categorize the underlying causes of errors. Figure \ref{fig:error_pie_chart} presents the results, and representative failure cases for each category are detailed in Appendix \ref{Case_study}. At a high level, the image-only input yields the lowest total error count, followed by the combined input, while the OCR-only input exhibits the highest error rate. We categorize errors into three main types: text misinterpretation (Error \textbf{A}), which involves challenges in aligning extracted information with the structured information; image-to-text extraction issues (Error \textbf{B}), which assess how well MLLMs understand textual content from images; and OCR schema ambiguity issues (Error \textbf{C}), which stem from inaccuracies in text recognition and confusion in document schema description.

We observe that image-to-text extraction errors are relatively high for the image-only setting but lower when OCR is included. This is expected, as our OCR system provides high transcription accuracy, whereas raw MLLMs may naturally introduce text-recognition errors. However, schema-ambiguity errors are notably reduced with image-only input. A likely explanation is that the built-in vision encoder integrates more effectively with the text encoder–decoder and captures page layout and document structure more faithfully, resulting in fewer overall mistakes. Nonetheless, there remains substantial room for improvement.
Motivated by these, we apply several enhancements to further improve performance:
\begin{itemize}
    \item Prompt Optimization: Introducing explicit emphasis and reasoning cues to encourage a more thoughtful generation.
    \item Format Refinement: Strengthening format constraints to reduce output inconsistencies.
    \item Schema Adjustment: Clarifying schema descriptions to minimize ambiguity.
\end{itemize}

Using these improvements, we performed a follow-up comparison experiment using a refined prompt template (details in Appendix \ref{new prompt}) for the input of only images. As shown in Table \ref{tab:OP_results}, the results show a further boost in performance, with the mean score increasing from 76.8\% to 78.9\%, which surpasses both the OCR-only and combined inputs. This promising result further validates the feasibility and effectiveness of the image-only approach in document information extraction.

\begin{table}
\centering
\resizebox{0.66\linewidth}{!}{%
\begin{tabular}{ccc}
\toprule
\multicolumn{1}{c}{\begin{tabular}[c]{@{}c@{}}\textbf{Google}\\ Gemini 1.5 Pro\end{tabular}} & \textbf{Initial} & \textbf{Final} \\
\midrule
Dataset C1 & 87.3 & \textbf{89.1} \\
Dataset C2 & 66.4 & \textbf{68.6} \\
Mean & 76.8 & \textbf{78.9} \\
\bottomrule
\end{tabular}
}
\caption{Performance results for the optimized prompt template with image-only input.}
\label{tab:OP_results}
\end{table}

\section{Conclusion}
In summary, we conducted a comprehensive benchmarking study on two internal document information extraction datasets, evaluating three distinct input modalities: OCR-only, image-only, and image+OCR. In addition, we perform an automatic error analysis in failure cases. Our findings reveal that powerful MLLMs can achieve competitive performance with image-only input, suggesting that OCR is not necessary in some cases. Furthermore, our automated error analysis helps developers identify common error patterns. Based on these, we demonstrate how well-designed schemas, exemplars, and instructions can further improve MLLM performance. We believe that these findings offer valuable insight to advance research in document information extraction.

\section*{Limitations}

Despite the promising results, our current approach has several limitations.
First, we did not systematically validate the effectiveness of few-shot learning.
Second, incorporating chain-of-thought (CoT) or a self-reflection mechanism could potentially further improve model performance, but this was not explored in our current setup due to the resource constraint.
Finally, our error analysis framework could further benefit from enhanced reasoning capabilities by integrating a reasoning model, such as O1~\cite{jaech2024openai} or DeepSeek R1~\cite{guo2025deepseek}. Exploring the use of such reasoning-centric models represents a direction for future work.

\bibliography{custom}

@inproceedings{huang2022layoutlmv3,
  title={{LayoutLMv3}: Pre-training for document ai with unified text and image masking},
  author={Huang, Yupan and Lv, Tengchao and Cui, Lei and Lu, Yutong and Wei, Furu},
  booktitle={Proceedings of the 30th ACM international conference on multimedia},
  pages={4083--4091},
  year={2022}
}

@article{katti2018chargrid,
  title={Chargrid: Towards understanding 2d documents},
  author={Katti, Anoop Raveendra and Reisswig, Christian and Guder, Cordula and Brarda, Sebastian and Bickel, Steffen and H{\"o}hne, Johannes and Faddoul, Jean Baptiste},
  journal={arXiv preprint arXiv:1809.08799},
  year={2018}
}

@article{bai2023qwen,
  title={Qwen technical report},
  author={Bai, Jinze and Bai, Shuai and Chu, Yunfei and Cui, Zeyu and Dang, Kai and Deng, Xiaodong and Fan, Yang and Ge, Wenbin and Han, Yu and Huang, Fei and others},
  journal={arXiv preprint arXiv:2309.16609},
  year={2023}
}

@article{team2023gemini,
  title={Gemini: a family of highly capable multimodal models},
  author={Team, Gemini and Anil, Rohan and Borgeaud, Sebastian and Alayrac, Jean-Baptiste and Yu, Jiahui and Soricut, Radu and Schalkwyk, Johan and Dai, Andrew M and Hauth, Anja and Millican, Katie and others},
  journal={arXiv preprint arXiv:2312.11805},
  year={2023}
}

@inproceedings{kim2022ocr,
  title={{OCR}-free document understanding transformer},
  author={Kim, Geewook and Hong, Teakgyu and Yim, Moonbin and Nam, JeongYeon and Park, Jinyoung and Yim, Jinyeong and Hwang, Wonseok and Yun, Sangdoo and Han, Dongyoon and Park, Seunghyun},
  booktitle={European Conference on Computer Vision},
  pages={498--517},
  year={2022},
  organization={Springer}
}

@article{ye2023mplug,
  title={{mPLUG-DocOwl}: Modularized multimodal large language model for document understanding},
  author={Ye, Jiabo and Hu, Anwen and Xu, Haiyang and Ye, Qinghao and Yan, Ming and Dan, Yuhao and Zhao, Chenlin and Xu, Guohai and Li, Chenliang and Tian, Junfeng and others},
  journal={arXiv preprint arXiv:2307.02499},
  year={2023}
}

@inproceedings{chen2024automatic,
  title={Automatic root cause analysis via large language models for cloud incidents},
  author={Chen, Yinfang and Xie, Huaibing and Ma, Minghua and Kang, Yu and Gao, Xin and Shi, Liu and Cao, Yunjie and Gao, Xuedong and Fan, Hao and Wen, Ming and others},
  booktitle={Proceedings of the Nineteenth European Conference on Computer Systems},
  pages={674--688},
  year={2024}
}

@article{liu2024textmonkey,
  title={{TextMonkey}: An ocr-free large multimodal model for understanding document},
  author={Liu, Yuliang and Yang, Biao and Liu, Qiang and Li, Zhang and Ma, Zhiyin and Zhang, Shuo and Bai, Xiang},
  journal={arXiv preprint arXiv:2403.04473},
  year={2024}
}

@inproceedings{wang2023vrdu,
  title={{VRDU}: A benchmark for visually-rich document understanding},
  author={Wang, Zilong and Zhou, Yichao and Wei, Wei and Lee, Chen-Yu and Tata, Sandeep},
  booktitle={Proceedings of the 29th ACM SIGKDD Conference on Knowledge Discovery and Data Mining},
  pages={5184--5193},
  year={2023}
}

@article{kaplan2020scaling,
  title={Scaling laws for neural language models},
  author={Kaplan, Jared and McCandlish, Sam and Henighan, Tom and Brown, Tom B and Chess, Benjamin and Child, Rewon and Gray, Scott and Radford, Alec and Wu, Jeffrey and Amodei, Dario},
  journal={arXiv preprint arXiv:2001.08361},
  year={2020}
}

@article{lee2024vhelm,
  title={{VHELM}: A holistic evaluation of vision language models},
  author={Lee, Tony and Tu, Haoqin and Wong, Chi Heem and Zheng, Wenhao and Zhou, Yiyang and Mai, Yifan and Roberts, Josselin and Yasunaga, Michihiro and Yao, Huaxiu and Xie, Cihang and Liang, Percy},
  journal={Advances in Neural Information Processing Systems},
  volume={37},
  pages={140632--140666},
  year={2024}
}

@misc{gartner-reviews,
    author= {Gartner},
    title= {Intelligent Document Processing Solutions Reviews and Ratings},
    publisher={Gartner},
    howpublished={\url{https://www.gartner.com/reviews/market/intelligent-document-processing-solutions}},
}

@misc{mistralocr,
    author= {MistralAI},
    title= {Mistral OCR Technique Report},
    howpublished={\url{https://mistral.ai/news/mistral-ocr}},
}

@article{hurst2024gpt,
  title={Gpt-4o system card},
  author={Hurst, Aaron and Lerer, Adam and Goucher, Adam P and Perelman, Adam and Ramesh, Aditya and Clark, Aidan and Ostrow, AJ and Welihinda, Akila and Hayes, Alan and Radford, Alec and others},
  journal={arXiv preprint arXiv:2410.21276},
  year={2024}
}

@inproceedings{ouyang2025omnidocbench,
  title={Omnidocbench: Benchmarking diverse pdf document parsing with comprehensive annotations},
  author={Ouyang, Linke and Qu, Yuan and Zhou, Hongbin and Zhu, Jiawei and Zhang, Rui and Lin, Qunshu and Wang, Bin and Zhao, Zhiyuan and Jiang, Man and Zhao, Xiaomeng and others},
  booktitle={Proceedings of the Computer Vision and Pattern Recognition Conference},
  pages={24838--24848},
  year={2025}
}

@article{guo2025deepseek,
  title={Deepseek-r1: Incentivizing reasoning capability in llms via reinforcement learning},
  author={Guo, Daya and Yang, Dejian and Zhang, Haowei and Song, Junxiao and Zhang, Ruoyu and Xu, Runxin and Zhu, Qihao and Ma, Shirong and Wang, Peiyi and Bi, Xiao and others},
  journal={arXiv preprint arXiv:2501.12948},
  year={2025}
}

@article{jaech2024openai,
  title={Openai o1 system card},
  author={Jaech, Aaron and Kalai, Adam and Lerer, Adam and Richardson, Adam and El-Kishky, Ahmed and Low, Aiden and Helyar, Alec and Madry, Aleksander and Beutel, Alex and Carney, Alex and others},
  journal={arXiv preprint arXiv:2412.16720},
  year={2024}
}

@article{team2024gemini,
  title={Gemini 1.5: Unlocking multimodal understanding across millions of tokens of context},
  author={Team, Gemini and Georgiev, Petko and Lei, Ving Ian and Burnell, Ryan and Bai, Libin and Gulati, Anmol and Tanzer, Garrett and Vincent, Damien and Pan, Zhufeng and Wang, Shibo and others},
  journal={arXiv preprint arXiv:2403.05530},
  year={2024}
}

@article{intelligence2024amazon,
  title={The amazon nova family of models: Technical report and model card},
  author={Intelligence, Amazon Artificial General},
  year={2024}
}

\appendix

\newpage

\section{Details in Evaluation Pipeline}
\label{details in eval}
We use the following prompt template in our original evaluation pipeline:

\begin{center}
\begin{tcolorbox}
[title=Prompt Template:, fit basedim=10pt, fonttitle=\sffamily\bfseries\small, width=\linewidth, fontupper=\footnotesize]

You are a warehouse manager receiving a delivery. As an expert, you go through the attached delivery note and carefully extract the data that you require to receive the shipped goods and process them in your ERP system. So it is important to focus on the actually received goods and quantities.\\

The document may be in English, German or any other language. Some of the fields that you need may be indicated by abbreviations in the language of the document. It is important that you carefully extract the information and that you only retrieve information actually on the document. If you have any doubts on a field, skip the field.\\

Instructions: \{format instructions\}. \\
\{document schema\}. \\

Return date fields in YYYY-MM-DD format.\\
For country and currency use ISO format.\\
Do not include the schema in the answer.\\
Return missing values as empty string.\\
Always return valid json and don't wrap you response in backticks!\\
Do not include a comma before the closing curly bracket.\\

Here is the document: \{OCR extracted content\}\\

Here is the image:

\end{tcolorbox}
\end{center}

The response format is like below:

\begin{center}
\begin{tcolorbox}[fit basedim=10pt, fonttitle=\sffamily\bfseries\small,colback=green!5!white,colframe=green!50!black,title=Response Example:, width=\linewidth, fontupper=\footnotesize]
\tcbfontsize{0.65}\texttt{\{\\
  "deliveryDate": [""],\\
  "deliveryNoteNumber": ["ID"],\\
  "documentDate": ["YYYY-MM-DD"],\\
  "purchaseOrderNumber": [""],\\
  "supplierId": [""],\\
  "lineItems": [\\
    \{\\
      "lineItem.customerMaterialNumber": "",\\
      "lineItem.itemNumber": "1",\\
      "lineItem.purchaseOrderItemNumber": "",\\
      "lineItem.purchaseOrderNumber": "",\\
      "lineItem.quantity": "QUANTITY",\\
      "lineItem.supplierMaterialNumber": "MATERIAL CODE",\\
      "lineItem.unitOfMeasure": ""\\
    \},\\
    ...\\
  ]\\
\}}
\end{tcolorbox}
\end{center}

\section{Dataset Statistics}
\label{appendix:data distribution}
A summary of our dataset statistics is provided in Table~\ref{tab:dataset_stats}:
\newpage
\begin{table*}[t]
\centering
\renewcommand{\arraystretch}{1.15}
\resizebox{\textwidth}{!}{%
\begin{tabular}{|p{2cm}|p{2.5cm}|p{2.5cm}|p{5cm}|p{6cm}|p{6cm}|}
\hline
\textbf{Dataset} & \textbf{Approx. Doc Count} & \textbf{Avg. Word Density} & \textbf{Page Language Distribution} & \textbf{Document Currencies} & \textbf{Document Countries} \\
\hline
C1 + C2 & Around 1,000 & High density (financial tabular + semi-structured text, $\sim$150--400 words per page) & English ($\sim$200), Spanish ($\sim$150), French ($\sim$100), Italian ($\sim$80), German ($\sim$90), Romanian ($\sim$20), Slovak ($\sim$10), Hungarian ($\sim$10), Portuguese ($\sim$10), Mixed/Unknown ($\sim$150), Other ($<$10 each) & Euro ($\sim$70), Indian Rupee ($\sim$70), US Dollar ($\sim$50), British Pound ($\sim$30), Generic/Masked ($\sim$500), Chinese Yuan ($\sim$10), UAE Dirham ($\sim$10), Indonesian Rupiah ($\sim$10), Swiss Franc ($\sim$10), Vietnamese Dong ($\sim$5), Malaysian Ringgit ($\sim$5), Saudi Riyal ($\sim$5), Venezuelan Bolívar ($\sim$5), Australian Dollar ($\sim$5), Philippine Peso ($\sim$5), South Korean Won ($<$5), South African Rand ($<$5), Singapore Dollar ($<$5), Moroccan Dirham ($<$5), New Zealand Dollar ($<$5), Bolivian Boliviano ($<$5), Canadian Dollar ($<$5), Azerbaijani Manat ($<$5), Turkish Lira ($<$5), Hungarian Forint ($<$5), Danish Krone ($<$5), Null/Unspecified ($\sim$20) & Spain ($\sim$120), Romania ($\sim$80), France ($\sim$80), Italy ($\sim$80), Germany ($\sim$90), India ($\sim$70), Netherlands ($\sim$70), US ($\sim$40), UK ($\sim$30), UAE ($\sim$10), China ($\sim$10), Indonesia ($\sim$10), Venezuela ($\sim$10), Saudi Arabia ($\sim$10), Ireland ($\sim$10), Austria ($\sim$5), Switzerland ($\sim$5), Denmark ($\sim$5), Slovakia ($\sim$5), Vietnam ($\sim$5), Malaysia ($\sim$5), Portugal ($\sim$5), Hungary ($\sim$5), Australia ($\sim$5), Philippines ($\sim$5), South Korea ($<$5), South Africa ($<$5), Singapore ($<$5), Morocco ($<$5), Peru ($<$5), New Zealand ($<$5), Bolivia ($<$5), Canada ($<$5), Azerbaijan ($<$5), Turkey ($<$5), Null/Unspecified ($\sim$10) \\
\hline
\end{tabular}%
}
\caption{Dataset characteristics: document counts, density, language distribution, currencies, and country coverage.}
\label{tab:dataset_stats}
\end{table*}

\section{Why some MLLMs perform even better with only image as input?}
\label{appendix:explanation}
Empirically, high-capacity models (e.g., Gemini variants, Nova Pro) often match or even surpass multimodal inputs when given image-only inputs. We identify two main drivers behind this pattern.

First, at the mechanistic level, web-scale pretraining equips these MLLMs with strong implicit OCR: visual tokenizers and 2D attention layers can recover glyphs, reading order, and layout hierarchies directly from images. This preserves typographic and spatial cues that external OCR systems may distort or lose. Consistent with this, our error analysis shows that OCR-only inputs are dominated by schema-ambiguity errors, whereas image-only inputs yield fewer total errors.

Second, scaling amplifies these advantages. As model capacity increases and instruction tuning improves, MLLMs internalize increasingly robust text recognition and layout-aware reasoning. This narrows—and occasionally reverses—the expected multimodal advantage. For example, as shown in Figure~\ref{fig:scaling_perf}, Gemini 2.0 Flash-Lite’s image-only configuration slightly surpasses its image+OCR setting.

\section{Failure Case Study}
\label{Case_study}
\subsection{Text misinterpretation}

\subsubsection*{Example 1}
\begin{figure}[t]
    \centering 
    \includegraphics[width=\linewidth]{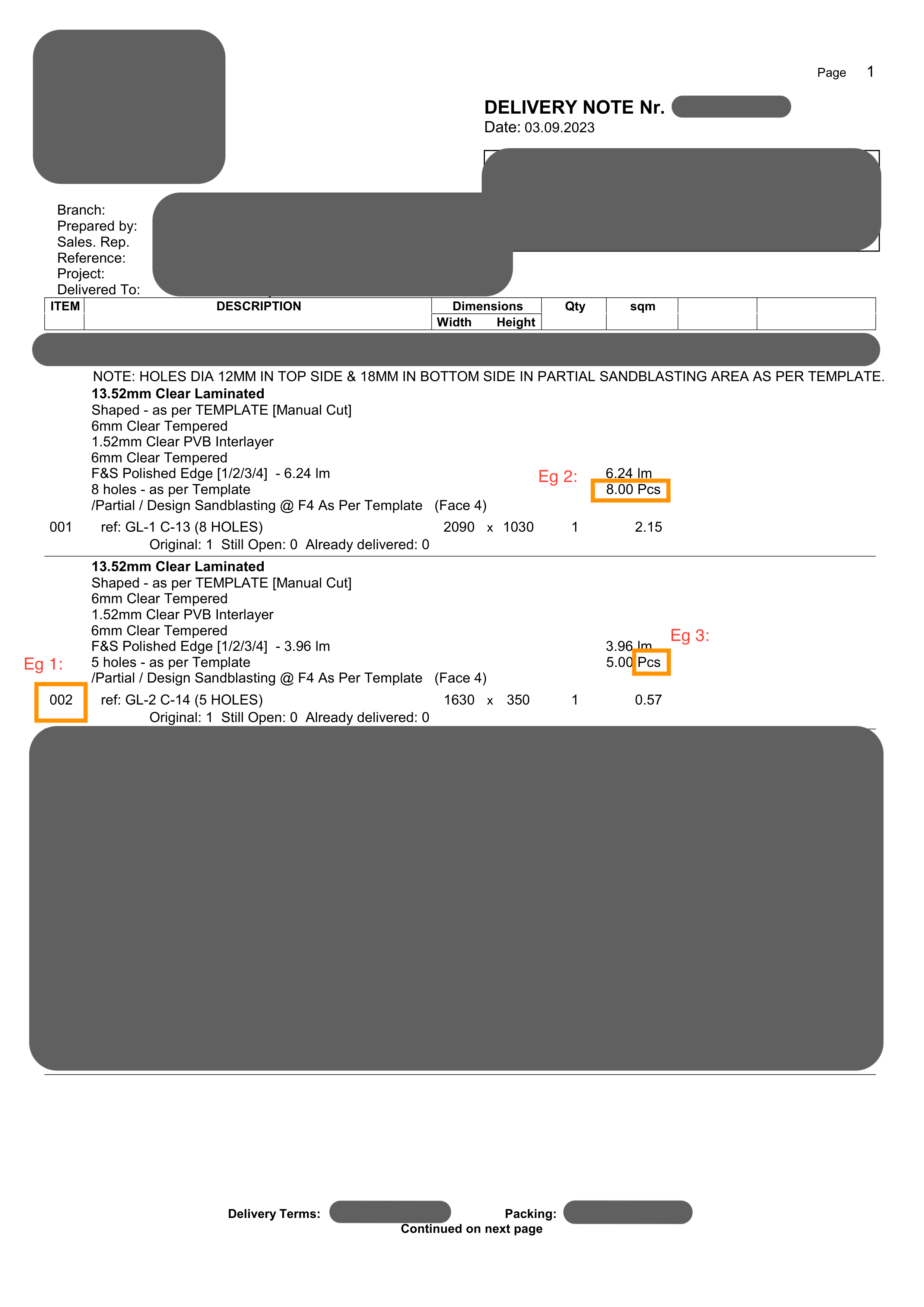}
    \caption{The corresponding image(cropped and censored) for example 1,2 and 3.}
    \label{fig:example1}
\end{figure}

For the data entry "lineItem.itemNumber", the ground truth specifies the item number as "2" while the prediction erroneously records it as "002". The cause analysis indicates that this mistake is likely from a misreading or misunderstanding of the given text format. The item number as shown in Figure \ref{fig:example1} is "002" confirms the correct OCR extraction. This suggests that the error is due to omission in the interpretation of the format guideline. 
\begin{center}
\begin{tcolorbox}
[title=Example 1:, fit basedim=10pt, fonttitle=\sffamily\bfseries\small, width=\linewidth, fontupper=\footnotesize]
\indent \textbf{Data entry:} \texttt{"lineItem.itemNumber"} \\
\indent \textbf{Ground truth:} \(\left[ \text{"2"} \right] \) \\
\indent \textbf{Prediction:} \texttt{"002"} \\
\indent \textbf{Cause:} \textit{"Error due to misreading or misunderstanding the text format"}
\end{tcolorbox}
\end{center}

\begin{center}
\begin{tcolorbox}
[title=Example 2:, fit basedim=10pt, fonttitle=\sffamily\bfseries\small, width=\linewidth, fontupper=\footnotesize]
\indent \indent \textbf{Data entry:} \texttt{"lineItem.quantity"} \\
\indent \textbf{Ground truth:} \(\left[ \text{"8.00} \right] \) \\
\indent \textbf{Prediction:} \texttt{"1"} \\
\indent \textbf{Cause:} \textit{"Error due to incorrect quantity extraction"}

\end{tcolorbox}
\end{center}

\subsubsection*{Example 2}
For the data entry "lineItem.quantity", the ground truth specifies that the quantity should be "8.00", but the prediction inaccurately records it as "1". It is reasoned that this discrepancy arises from an error in the extraction process, where the quantity is incorrectly interpreted or extracted. The model does not capture "8.00Pcs" from the table in Figure \ref{fig:example1} and correctly identifies it as the quantity attribute, suggesting a text misinterpretation problem.

\subsubsection*{Example 3}
Following Example 2, the model fails to identify "Pcs" in "8.00 Pcs" as the unit of measure. Instead, the prediction is "Im". This error implies a misinterpretation of abbreviations during the data extraction process. 
\begin{center}
\begin{tcolorbox}
[title=Example 3:, fit basedim=10pt, fonttitle=\sffamily\bfseries\small, width=\linewidth, fontupper=\footnotesize]
\indent \indent \textbf{Data entry:} \texttt{"lineItem.unitOfMeasure"} \\
\indent \textbf{Ground truth:} \(\left[ \text{"Pcs"} \right] \) \\
\indent \textbf{Prediction:} \texttt{"Im"} \\
\indent \textbf{Cause:} \textit{"Error due to misinterpretation of abbreviations"}

\end{tcolorbox}
\end{center}

\subsection{Image-to-text extraction issue}

\subsubsection*{Example 4}

\begin{figure}
    \centering
    \includegraphics[width=\linewidth]{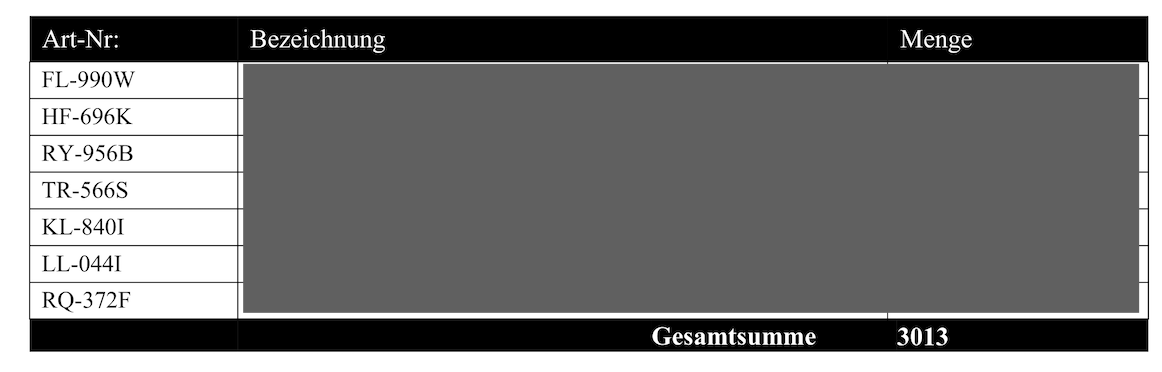}
    \caption{The corresponding image(cropped and censored) for example 4.}
    \label{fig:example4}
\end{figure}

Regarding with the data entry "lineItem.supplierMaterialNumber", the ground truth specifies "KL‑840I" whereas the prediction is "KL‑8401". The cause analysis suggests that the error arises from visual similarity between the character "I" and the digit "1" in the document image, as shown in Figure~\ref{fig:example4}. As the model performs direct image-to-text extraction without explicit OCR segmentation, it misinterpreted the final character due to font style, resolution, or noise, replacing the uppercase "I" with the numeral "1".

\begin{center}
\begin{tcolorbox}
[title=Example 4:, fit basedim=10pt, fonttitle=\sffamily\bfseries\small, width=\linewidth, fontupper=\footnotesize]
\indent \textbf{Data entry:} \texttt{"lineItem.supplierMaterialNumber"} \\
\indent \textbf{Ground truth:} \(\left[ \text{"KL-840I"} \right] \) \\
\indent \textbf{Prediction:} \texttt{"KL-8401"} \\
\indent \textbf{Cause:} \textit{"The model misinterpreted the quantity field as the item number due to their close proximity within the document."}
\end{tcolorbox}
\end{center}

\subsubsection*{Example 5}

\begin{figure}
    \centering
    \includegraphics[width=\linewidth]{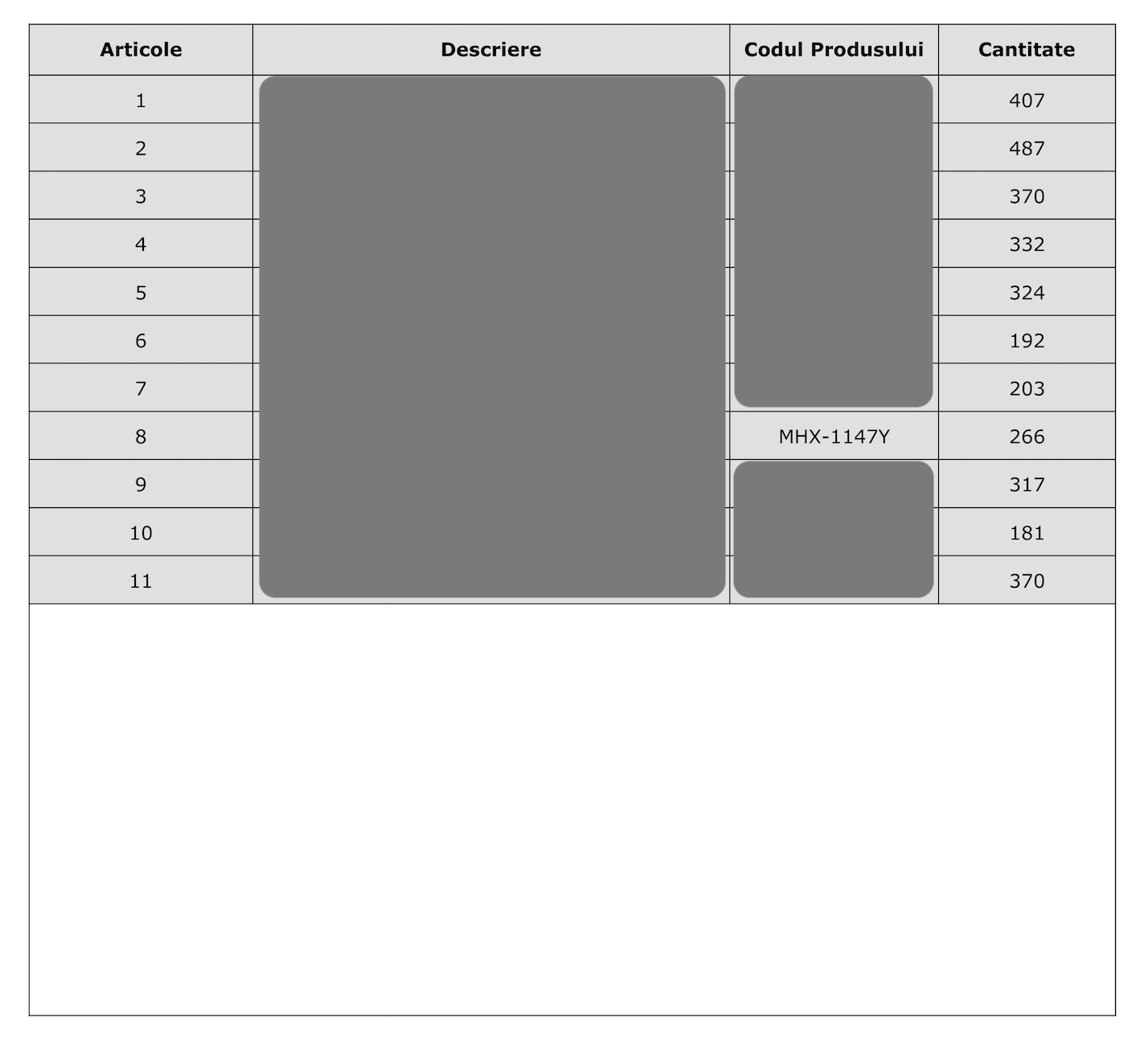}
    \caption{The corresponding image(cropped and censored) for example 5.}
    \label{fig:example5}
\end{figure}

As shown in Figure \ref{fig:example5}, for the data entry "lineItem.supplierMaterialNumber", the ground truth specifies "MHX-1147Y", whereas the prediction incorrectly records it as "$\mathrm{\textit{MHX}}$-1147Y". This error stems from the misinterpretation of the character "X" as the Greek letter "$\mathrm{\textit{X}}$" (Chi), due to their visual similarity.
\begin{center}
\begin{tcolorbox}
[title=Example 5:, fit basedim=10pt, fonttitle=\sffamily\bfseries\small, width=\linewidth, fontupper=\footnotesize]
\indent \textbf{Data entry:} \texttt{"lineItem.supplierMaterialNumber"} \\
\indent \textbf{Ground truth:} \(\left[ \text{"MHX-1147Y"} \right] \) \\
\indent \textbf{Prediction:} \texttt{"\textbackslash u039c\textbackslash u0397\textbackslash u03a7-1147Y"} \\
\indent \textbf{Cause:} \textit{"The character 'X' was misinterpreted as the Greek letter '$\mathrm{X}$'."}
\end{tcolorbox}
\end{center}

\begin{figure}
    \centering
    \includegraphics[width=\linewidth]{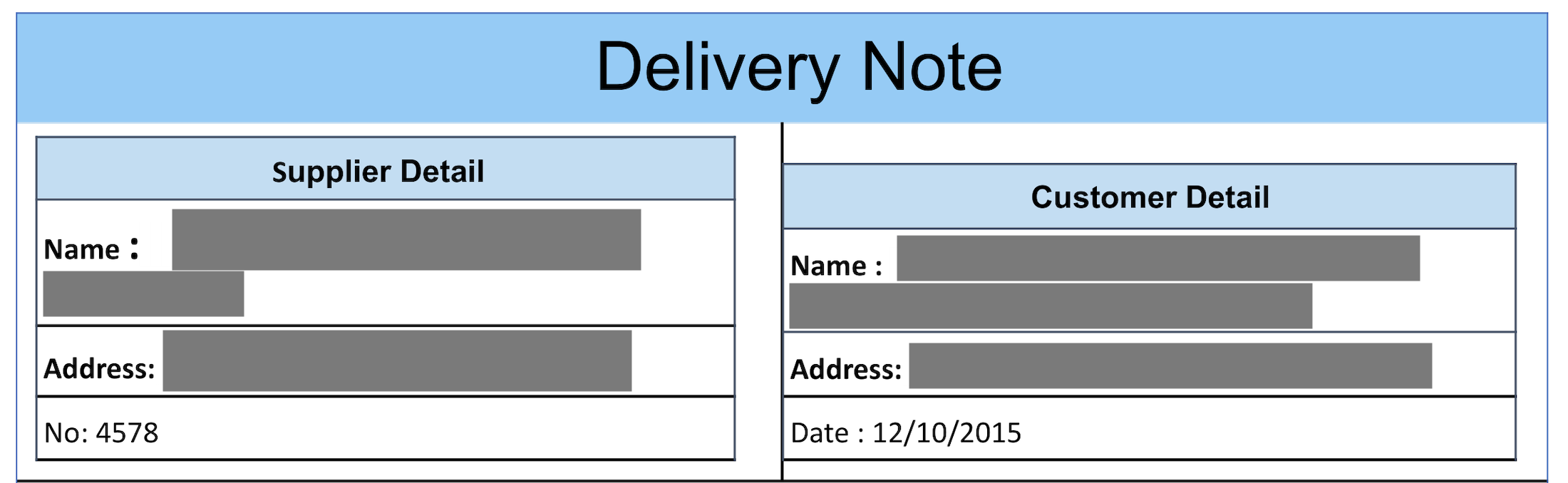}
    \caption{The corresponding image(cropped and censored) for example 6.}
    \label{fig:example6}
\end{figure}

\subsubsection*{Example 6}
For the data entry "deliveryNoteNumber", the ground truth indicates "4578" but the prediction yields an empty result. The cause analysis shows that the field is not recognized in the image text. In Figure \ref{fig:example6}, the ground truth "4578" appears under "Supplier Detail" rather than being explicitly labelled as "deliveryNoteNumber", presenting a challenge for the extraction model in terms of high-level layout comprehension and reasoning. 
\begin{center}
\begin{tcolorbox}
[title=Example 6:, fit basedim=10pt, fonttitle=\sffamily\bfseries\small, width=\linewidth, fontupper=\footnotesize]
\indent \textbf{Data entry:} \texttt{"deliveryNoteNumber"} \\
\indent \textbf{Ground truth:} \(\left[ \text{"4578"} \right] \) \\
\indent \textbf{Prediction:} \texttt{""} \\
\indent \textbf{Cause:} \textit{"Prediction was empty because the field was not explicitly recognized in the image text."}
\end{tcolorbox}
\end{center}

\subsection{OCR schema ambiguity}

\subsubsection*{Example 7}
\begin{figure}
    \centering
    \includegraphics[width=\linewidth]{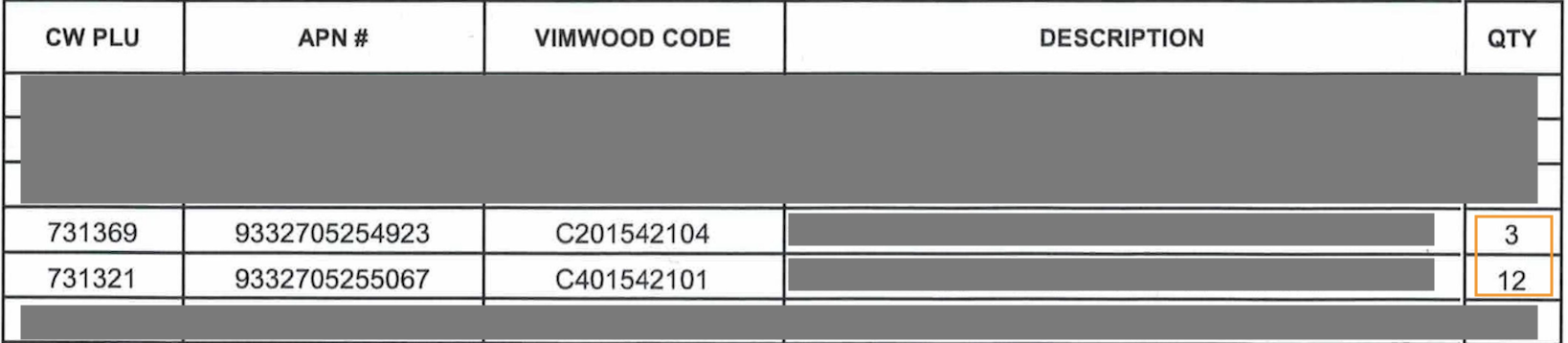}
    \caption{The corresponding image(cropped and censored) for example 7.}
    \label{fig:example7}
\end{figure}

For the data entry "lineItem.quantity", the ground truth specifies "3" whereas the prediction inaccurately states "12" The cause analysis suggests that the error is due to incorrect logic or misalignment in OCR. In Figure \ref{fig:example7}, both "3" and "12" are located within the quantity column, but they appear in different rows. OCR misalignment or incomplete structured data led the prediction to mistakenly extract "12" from a neighboring row, rather than the correct value "3".
\begin{center}
\begin{tcolorbox}
[title=Example 7:, fit basedim=10pt, fonttitle=\sffamily\bfseries\small, width=\linewidth, fontupper=\footnotesize]
\indent \textbf{Data entry:} \texttt{"lineItem.quantity"} \\
\indent \textbf{Ground truth:} \(\left[ \text{"3"} \right] \) \\
\indent \textbf{Prediction:} \texttt{"12"} \\
\indent \textbf{Cause:} \textit{"Incorrect logic or misalignment in OCR could cause quantity mismatch."}
\end{tcolorbox}
\end{center}

\subsubsection*{Example 8 \& 9}

\begin{figure}
    \centering
    \includegraphics[width=\linewidth]{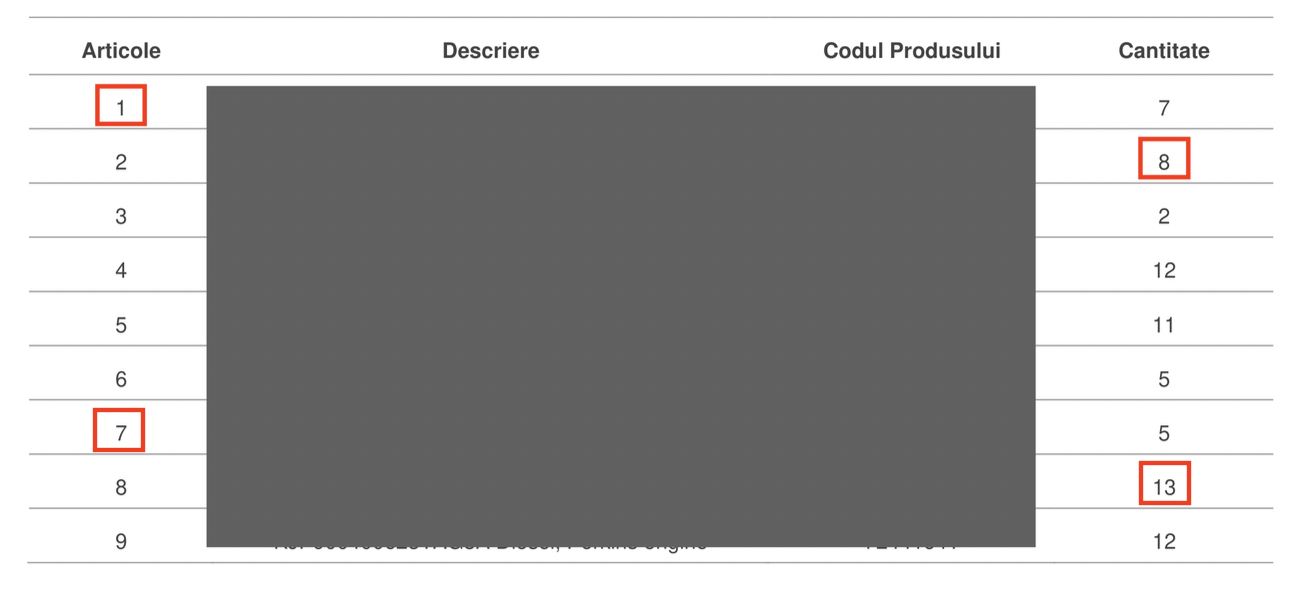}
    \caption{The corresponding image(cropped and censored) for example 8 and 9.}
    \label{fig:example8}
\end{figure}

For the data entries "lineItem.itemNumber" and "lineItem.quantity", the ground truth specifies "1" and "13", whereas the predictions are "8" and "7", respectively. The cause analysis suggests that the error results from OCR extracting both fields as adjacent tokens without clear separation or labeling. In the OCR output, the item number and quantity values appear consecutively in a single text segment or without distinct bounding boxes. As a result, when the LLM processes this unstructured or ambiguously segmented text, it may confuse the associations between values and fields. In this case, the model likely misaligned the detected numbers, attributing "8" to the item number and "7" to the quantity, rather than correctly mapping "1" and "13". Figure~\ref{fig:example8} shows that the close spatial proximity of numeric fields contributed to this misinterpretation.

\begin{center}
\begin{tcolorbox}
[title=Example 8:, fit basedim=10pt, fonttitle=\sffamily\bfseries\small, width=\linewidth, fontupper=\footnotesize]
\indent \textbf{Data entry:} \texttt{"lineItem.itemNumber"} \\
\indent \textbf{Ground truth:} \(\left[ \text{"1"} \right] \) \\
\indent \textbf{Prediction:} \texttt{"8"} \\
\indent \textbf{Cause:} \textit{"The OCR data extracted the itemNumber and quantity as adjacent fields, which can lead to misinterpretation by the LLM."}
\end{tcolorbox}
\end{center}

\begin{center}
\begin{tcolorbox}
[title=Example 9:, fit basedim=10pt, fonttitle=\sffamily\bfseries\small, width=\linewidth, fontupper=\footnotesize]
\indent \textbf{Data entry:} \texttt{"lineItem.quantity"} \\
\indent \textbf{Ground truth:} \(\left[ \text{"13"} \right] \) \\
\indent \textbf{Prediction:} \texttt{"7"} \\
\indent \textbf{Cause:} \textit{"The OCR data extracted the itemNumber and quantity as adjacent fields, which can lead to misinterpretation by the LLM."}
\end{tcolorbox}
\end{center}

\newpage
\section{Refined Prompt Template}
\label{new prompt}

We cannot disclose the format instructions and document schema information. Therefore, we have omitted these two variables, but all other details for our refined prompt template are presented below:

\begin{center}
\begin{tcolorbox}
[title=Prompt Template for Image-only Input:, fit basedim=10pt, fonttitle=\sffamily\bfseries\small, width=\linewidth, fontupper=\footnotesize]
You are a warehouse manager receiving a delivery. As an expert, you will go through the attached delivery note and carefully extract the data required to receive the shipped goods and process them in your ERP system. Focus on the actually received goods and quantities.\\

The document may be in English, German, or any other language. Some fields may be indicated by abbreviations. Extract only the information present in the document. If you have doubts about a field, skip it.\\

Format instructions:  \{modified format instructions\}. \\
\{modified document schema\}. \\

Return date fields in YYYY-MM-DD format. For country and currency, use ISO format. Do not include the schema in the answer. Ensure that all fields are returned as valid values or empty strings (`""'), rather than null. If a field does not have a value, return it as an empty string.\\

Always return valid JSON and do not wrap your response in backticks! Ensure that the JSON structure is valid and does not contain any extra commas or brackets. Each object should be properly closed without trailing commas.\\

Be attentive to abbreviations and language variations in the document, and ensure that you extract the correct information based on context. Validate the JSON structure before returning the output, checking for any syntax errors. Accuracy in the extraction process is crucial, ensuring that all relevant details are captured accurately.\\

Emphasize the importance of accuracy in the extraction process and encourage the model to double-check its outputs against the provided schema. Pay special attention to context clues in the document to accurately extract and interpret abbreviations and language variations. Your output must reflect the exact information present in the document, as inaccuracies can lead to significant operational issues.\\

Here is the document image:

\end{tcolorbox}
\end{center}

\end{document}